\documentclass{article}

\usepackage[final]{tackling_climate_workshop_style}

\usepackage[utf8]{inputenc} 
\usepackage[T1]{fontenc}    
\usepackage{hyperref}       
\usepackage{url}            
\usepackage{booktabs}       
\usepackage{amsfonts}       
\usepackage{nicefrac}       
\usepackage{microtype}      
\usepackage{amsmath}
\usepackage{graphicx}
\graphicspath{ {../images/} }

\title{\textit{GraphTransformers} for Geospatial Forecasting  
of Hurricane Trajectories}

\author{%
  Pallavi Banerjee \\ 
  University of Washington\\
  \texttt{pallavib@uw.edu} \\
  \And
   Satyaki Chakraborty\\
   Carnegie Mellon University\\
  \texttt{schakra1@cs.cmu.edu} \\
}

\begin{document}

\maketitle

\begin{abstract}
  In this paper we introduce a novel framework for trajectory prediction of geospatial sequences using \textit{GraphTransformers}. When viewed across several sequences, we observed that a graph structure automatically emerges between different geospatial points that is often not taken into account for such sequence modeling tasks. We show that by leveraging this graph structure explicitly, geospatial trajectory prediction can be significantly improved. Our \textit{GraphTransformer} approach improves upon state-of-the-art Transformer based baseline significantly on HURDAT, a dataset where we are interested in predicting the trajectory of a hurricane on a 6 hourly basis. This helps inform evacuation efforts by narrowing down target location by $10$ to $20$ kilometers along both the north-south and east-west directions.
\end{abstract}

\section{Introduction}

Hurricanes, generally known as tropical cyclones are storm systems characterized by low pressure and high wind speed. These storms lead to physical and financial damage which have been on the rise due to climate change [1]. Identifying their trajectory will inform efficient rescue and evacuation methods by targeting the correct locations. This will reduce the damage to property and life. In this paper we are interested in the task of next location prediction  given the trajectory of a hurricane observed at a 6 hourly interval. 

\begin{figure}[!htb]
  \centering
  \includegraphics[scale=0.2]{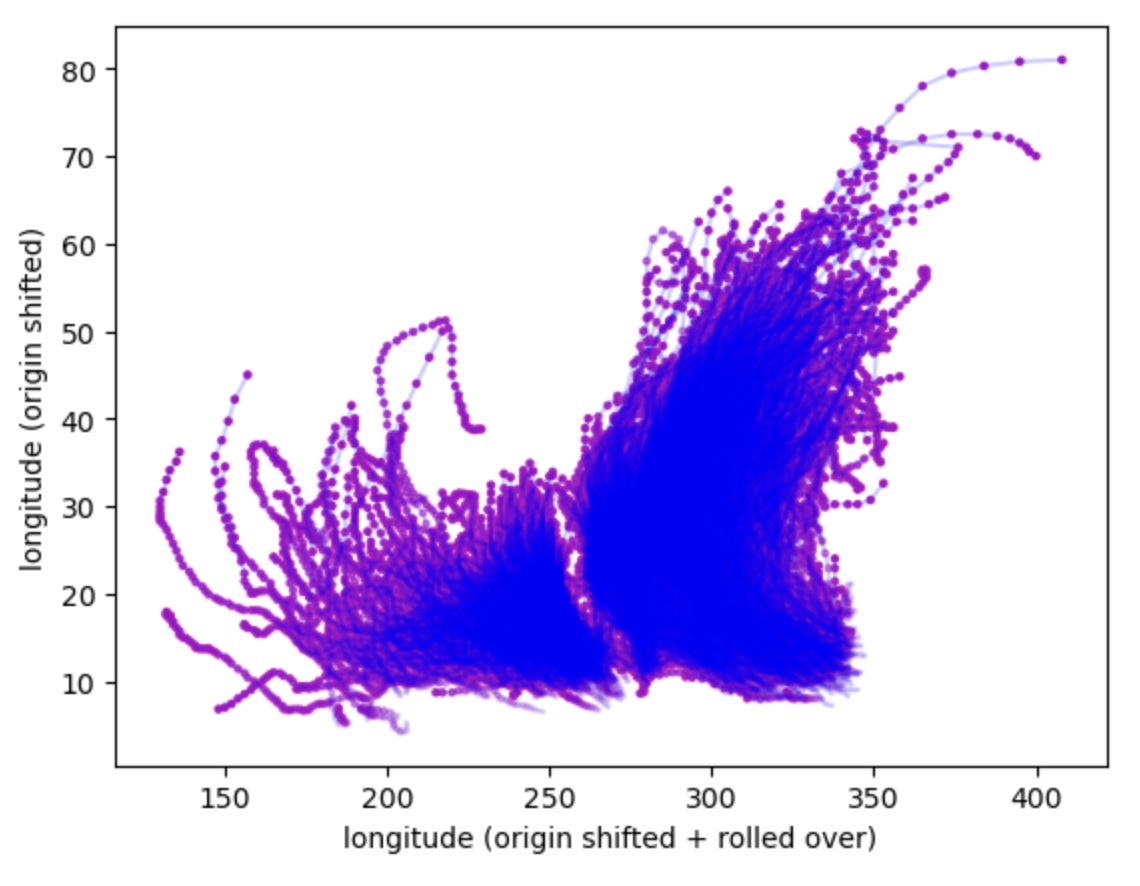}
  \caption{Storm trajectories over two centuries of data plotted across Northern America. Significant overlap and intersection in this graph suggest correlation between points across different trajectories, which is under-leveraged in prior work.}
\end{figure} 

When such geospatial sequences of extreme weather patterns are observed over a significant period of time an underlying directed graph structure emerges between the different spatial points. While traditional [2] and machine learning based studies [3, 4] for trajectory forecasting show promising results for this task, they typically rely on local context about the trajectory's past location to perform the prediction task. The global context that can be obtained in the form of a knowledge graph by looking across multiple such trajectories is often under-leveraged. In this paper we provide a general approach based on \textit{GraphTransformers} for modeling such geospatial sequences where the graph neural network captures the global context from a heuristically constructed knowledge graph and the transformer models the local context from past trajectory points.

\section{Related Work}

For hurricane trajectory prediction traditionally dynamic models, statistical models and statistical-dynamic models have been used. Dynamic models solve physical atmospheric equations using supercomputers whereas statistical models focus on the historic relation between storms features and location. Statistical-dynamic models blend both of the approaches together. Currently, neural network architectures have been employed to address similar tasks of trajectory estimation. Previous work [5]-[9] include using sophisticated deep learning architectures like recurrent neural networks to leverage the sequential storm trajectory, convolutional neural network to utilize satellite imagery data and multimodal approaches [10] combining different data sources. There are also several studies which discuss the application of a hybrid statistical-deep learning method in forecasting tasks [11]. However, none of these approaches have explicitly leveraged the underlying graph structure for hurricane prediction tasks. 

Graph neural networks, [12]-[24] are neural networks that are designed to model data that reside in a graph and are extensively used in a wide range of domains such as recommender systems, biomedical sciences, prediction, social media analysis [25]-[31].
Subsequently, transformers[32] have revolutionized the field of sequence modeling and have been used in different machine learning tasks in and outside of NLP such as classification, anomaly detection, translation, time series forecasting etc [33]-[36]. Graph Transformers are a fairly recent development which passes graph nodes as tokens and have been known to give promising results in prediction tasks [37]-[39]. In this paper we leverage the underlying graph structure of the hurricane trajectory using a GraphTransformer, where the transformer component models the trajectory sequential and the graph neural network is used for generating node embeddings 


\section{Approach}

\subsection{Dataset Preparation}
\label{dataprep}
HURDAT dataset contains location and weather information about $2864$ storm trajectory sequences collected from $1851$ to $2015$ in North America. We first preprocess the dataset to have training, validation and test splits roughly consisting $80 \%$, $10 \%$ and $10 \%$ of the trajectory sequences respectively. We use \textit{stratified sampling} to do the splits based on the length of the sequence to ensure that the splits cover storm sequences of varying lengths. We then create model input-output pairs ($X_i$, $Y_i$) from each sequence by randomly sampling a target location, $Y_i$ as the model target and using the sequence upto the target token (with a maximum sequence of $16$ tokens) as the corresponding model input, $X_i$. We do this to ensure there is no data leakage between the training and the test and validation sets. 

\subsection{Geospatial Graph Construction}
We observe by studying several such hurricane trajectory sequences that a natural directed graph structure emerges between different locations where the weight of an edge ($u \rightarrow v$) denotes the likelihood of the storm moving from location $u$ to location $v$ in the next time step. 
We construct this graph from the training set using a simple heuristic. We first create nodes corresponding to all latitude-longitude pairs in the training set with their latitude-longitude values being the node features. The latitude-longitude values are rounded off to the first decimal place. Then, for 
every $u_t$ in sequence $X_i$ we add directional edges with weight as follows where ever they are defined. 

\begin{gather*}
    W[u_{t - 1} \rightarrow u_t] \mathrel{+}= 1.0 \\ 
    W[u_{t - 2} \rightarrow u_t] \mathrel{+}= 0.5; 
    W[u_{t - 3} \rightarrow u_t] \mathrel{+}= 0.5 \\
    W[u_{t - 4} \rightarrow u_t] \mathrel{+}= 0.1; 
    W[u_{t - 5} \rightarrow u_t] \mathrel{+}= 0.1
\end{gather*}

\subsection{Feature Engineering}
The input features to the transformer at a given time step consist of the current location along with its local graph structure and weather conditions. For the graph features, we simply sample a subgraph from the $k$-distance ego-graph of the location node. Empirically, we found that using $k = 1$ achieves good results although increasing the value of $k$ can slightly improve the results at the cost of increase in memory and training time. All our experiments use $k = 1$ unless otherwise mentioned. During inference, it is not guaranteed that all location nodes will exist within our graph built from training set trajectories. For such cases we take the node representation of the nearest location in our training graph if it lies within an $L2$-distance of $0.75$ units otherwise we only use the latitude longitude features for the node. 

For the weather features which are numerical features denoting directional wind strengths, we simply standardize the values across the dataset to ensure that they follow the unit normal distribution. For ease of generalization, we transform all the latitude-longitude features (including the graph features and the target location) to the local co-ordinate frame with the starting location being the origin.

\begin{figure}[!htb]
  \centering
  \includegraphics[scale=0.6]{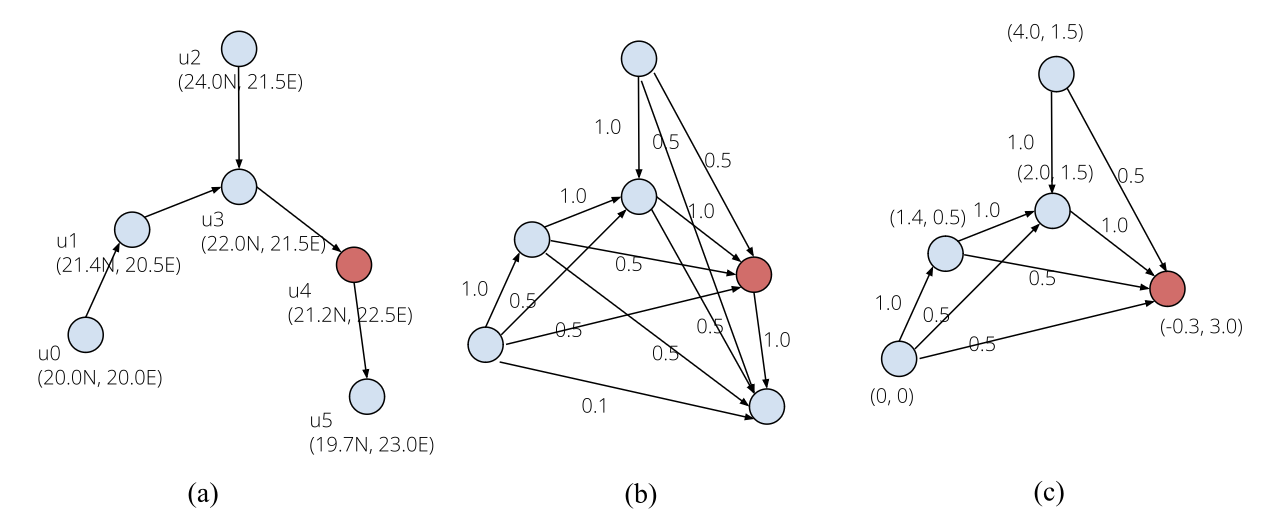}
  \caption{Example of the knowledge graph construction and neighborhood sampling. (a) A toy dataset of two trajectories $u_0 \rightarrow u_1 \rightarrow u_3 \rightarrow u_4 \rightarrow u_5$ and $u_2 \rightarrow u_3 \rightarrow u_4 \rightarrow u_5$ (b) Heuristically constructed  graph from the two trajectories (c) subgraph with node features sampled during training for node $u_4$}
\end{figure} 

\subsection{Architecture and Training}
\begin{figure}
  \centering
  \includegraphics[scale=0.45]{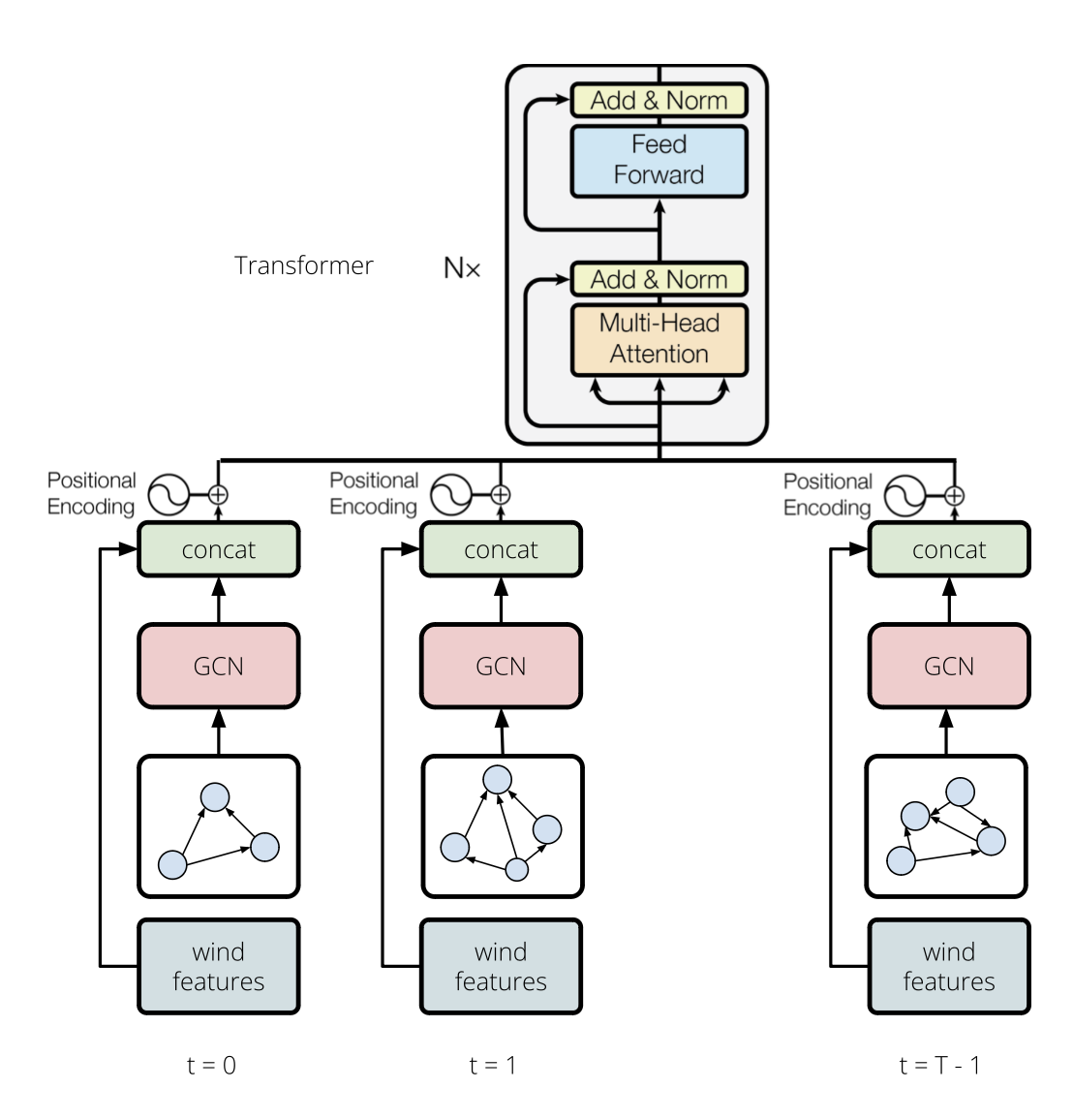}
  \caption{Our GraphTransformer model where the graph embedding from the spatial graph is concatenated with weather features and sent to the transformer for next location prediction}
  \label{fig:arch}
\end{figure}

Our graph transformer architecture Fig \ref{fig:arch} uses a GCN with $2$ graph convolution layers to encode graph features of the location node. The GCN takes node features which are $2$ dimensional since they denote the latitude longitude in the local co-ordinate frame. The output and intermediate representations of the GCN is $16$ dimensional. The output of the GCN is then concatenated with the standardized directional wind features and this concatenated input is then passed into a transformer encoder. The transformer encoder has $4$ layers of multihead self-attention with each self attention block having $4$ heads with an embedding dimension of $64$. At the output of the transformer encoder we take the representation of the final token from the sequence and use a linear layer to solve the regression problem. We use smoothL1 loss[40] function for the regression loss and adam optimizer with initial learning rate as $1e-4$. We train for $30$ epochs on a Tesla T4 GPU for all our experiments with early stopping enabled. 

\section{Evaluation and Results}
For evaluation we perform $5$-fold cross validation and show that graph transformers significantly outperform vanilla transformers in the task of next location prediction from a given trajectory. For fair evaluation we divide the test set into different buckets based on the length  of trajectory sequence that is provided as input to the model, compute mean absolute error for latitude longitude values predicted and then finally average across all the buckets. From table \ref{results-table} we show that graph transformers outperform vanilla transformers for all such buckets by significant margin.  


\begin{table}[!htb]
  \caption{Mean Absolute Error for different models}
  \label{results-table}
  \centering
  \begin{tabular}{lllll}
    \toprule
    Sequence Length     & \multicolumn{2}{c}{Transformer}  & \multicolumn{2}{c}{GraphTransformer (ours)} \\
    \midrule
      & Lat. & Long.  & Lat. & Long. \\
    \midrule
    0-5 & $0.62 ^\circ $ & $0.73 ^\circ$  & $0.47 ^\circ$ & $0.54 ^\circ$     \\
    6-10     & $0.53 ^\circ$ & $0.56 ^\circ$ & $0.36 ^\circ$ & $0.43 ^\circ$  \\
    11-15     & $0.48 ^\circ$ & $0.52 ^\circ$& $0.30 ^\circ$ & $0.32 ^\circ$  \\
    \midrule
    All sequences & $0.53 ^\circ$ & $0.60 ^\circ$ & \textbf{0.35} $^\circ$ & \textbf{0.41} $^\circ$ \\
    \bottomrule
  \end{tabular}
\end{table}

As shown in table \ref{results-table}, by leveraging geospatial graphs we improve upon SOTA sequential model approaches by $0.18^\circ$ in latitude and $0.19 ^\circ$ in longitude. This roughly translates to narrowing down the target location by around $20$ kilometers along N-S direction and around $10 - 15$ kilometers along E-W direction. Given that the US department of homeland security recommends [41] evacuation routes $30 - 80$ kilometers inland , a $10 - 20$ kilometer improvement in precision of a model that predicts target location on a 6 hourly basis, can hopefully save numerous lives.

\section{Conclusion and Future Work}

  In this paper we introduced a novel GraphTransformer based framework to predict the next affected location in the hurricane's trajectory. By improving the trajectory prediction task we hope to better inform risk mitigation and damage control efforts. In the future, for further performance enhancement we can expand the model to a multi-modal architecture by adding satellite imagery information, temperature, topography and precipitation data. We can also perform extensive hyperparameter tuning and network architecture search to further improve the performance. 

\section*{References}

\medskip

\small


[1] https://www.ncei.noaa.gov/access/billions/events/US/1980-2023?disasters[]=all-disasters

[2] Marks, F. D., DeMaria, M. 2003, Development of a Tropical Cyclone Rainfall Climatology and Persistence (R-CLIPER) Model

[3] Wang, D., Liu, B., Tan, P., and Liu, L. (2020). OMULET: Online Multi-Lead Time Location Prediction for Hurricane Trajectory Forecasting. Proceedings of the AAAI Conference on Artificial Intelligence, 34(01), 963–970. 

[4] Alemany, S., Beltran, J., Pérez, A. S., and Ganzfried, S. (2019). Predicting hurricane trajectories using a recurrent neural network. Proceedings of the AAAI Conference on Artificial Intelligence, 33(01), 468–475. 

[5] Fogarty, E. A., Elsner, J. B., Jagger, T. H., and Tsonis, A. A. (2008). Network analysis of U.S. hurricanes. In Springer eBooks (pp. 153–167).

[6] J. R. Rhome. Technical summary of the national hurricane center track and intensity models. Updated September, 12:2007, 2007.

[7] M. Moradi Kordmahalleh, M. Gorji Sefidmazgi, and A. Homaifar. A sparse recurrent neural network for trajectory prediction of Atlantic hurricanes. In Proceedings of the Genetic and Evolutionary Computation Conference 2016, pages 957–964, 2016.

[8] S. Alemany, J. Beltran, A. Perez, and S. Ganzfried. Predicting hurricane trajectories using a recurrent neural network. In Proceedings of the AAAI Conference on Artificial Intelligence, volume 33, pages 468–475, 2019.

[9] S. Gao, P. Zhao, B. Pan, Y. Li, M. Zhou, J. Xu, S. Zhong, and Z. Shi. A nowcasting model for the prediction of typhoon tracks based on a long short term memory neural network. Acta Oceanologica Sinica, 37:8–12, 2018.

[10] J. Lian, P. Dong, Y. Zhang, J. Pan, and K. Liu. A novel data-driven tropical cyclone track prediction model based on cnn and gru with multi-dimensional feature selection. IEEE Access, 2020.

[11] S. Giffard-Roisin, M. Yang, G. Charpiat, C. Kumler Bonfanti, B. Kégl, and C. Monteleoni. Tropical cyclone track forecasting using fused deep learning from aligned reanalysis data. Frontiers in Big Data, 3:1, 2020.

[12] Shiyang Li, Xiaoyong Jin, Yao Xuan, Xiyou Zhou, Wenhu Chen, Yu-Xiang Wang, and Xifeng Yan. Enhancing the locality and breaking the memory bottleneck of Transformers on time series forecasting. In NeurIPS, 2019.

[13] Tian Zhou, Ziqing Ma, Qingsong Wen, Xue Wang, Liang Sun, and Rong Jin. FEDformer: Frequency enhanced decomposed transformer for long-term series forecasting. In ICML, 2022.

[14] Wen, Q., Tian, Z., Zhang, C., Chen, W., Ma, Z., Yan, J., and Sun, L. (2022). Transformers in Time Series: A survey. arXiv (Cornell University). 

[15] F. Scarselli, M. Gori, A. C. Tsoi, M. Hagenbuchner, and G. Monfardini, “The graph neural network model,” IEEE Transactions on Neural Networks, vol. 20, no. 1, pp. 61–80, 2009.

[16] C. Gallicchio and A. Micheli, “Graph echo state networks,” in IJCNN. IEEE, 2010, pp. 1–8

[17] Y. Li, D. Tarlow, M. Brockschmidt, and R. Zemel, “Gated graph sequence neural networks,” in Proc. of ICLR, 2015.

[18] H. Dai, Z. Kozareva, B. Dai, A. Smola, and L. Song, “Learning steady states of iterative algorithms over graphs,” in Proc. of ICML, 2018, pp. 1114–1122.

[19] J. Bruna, W. Zaremba, A. Szlam, and Y. LeCun, “Spectral networks and locally connected networks on graphs,” in Proc. of ICLR, 2014.

[20] M. Henaff, J. Bruna, and Y. LeCun, “Deep convolutional networks on graph-structured data,” arXiv preprint arXiv:1506.05163, 2015.

[21] M. Defferrard, X. Bresson, and P. Vandergheynst, “Convolutional neural networks on graphs with fast localized spectral filtering,” in Proc. of NIPS, 2016, pp. 3844–3852.

[22] T. N. Kipf and M. Welling, “Semi-supervised classification with graph convolutional networks,” in Proc. of ICLR, 2017. 

[23] R. Levie, F. Monti, X. Bresson, and M. M. Bronstein, “Cayleynets: Graph convolutional neural networks with complex rational spectral filters,” IEEE Transactions on Signal Processing, vol. 67, no. 1, pp. 97–109, 2017.

[24] A. Micheli, “Neural network for graphs: A contextual constructive approach,” IEEE Transactions on Neural Networks, vol. 20, no. 3, pp. 498–511, 2009. 

[25] J. Atwood and D. Towsley, “Diffusion-convolutional neural networks,” in Proc. of NIPS, 2016, pp. 1993–2001.

[26] M. Niepert, M. Ahmed, and K. Kutzkov, “Learning convolutional neural networks for graphs,” in Proc. of ICML, 2016, pp. 2014–2023. [27] J. Gilmer, S. S. Schoeholz, P. F. Riley, O. Vinyals, and G. E. Dahl, “Neural message passing"

[27] Kipf, T. N. and Welling, M. Semi-supervised classification with graph convolutional networks. In ICLR, 2017.

[28] Ying, R., He, R., Chen, K., Eksombatchai, P., Hamilton, W. L., and Leskovec, J. Graph convolutional neural networks for web-scale recommender systems. In KDD, pp. 974–983. ACM, 2018.

[29] M. Mudigonda, S. Kim, A. Mahesh, S. Kahou, K. Kashinath, D. Williams, V. Michalski, T. O’Brien, and M. Prabhat. Segmenting and tracking extreme climate events using neural networks. 2017.

[30] Boussioux, L., Zeng, C., Guénais, T., and Bertsimas, D. (2022). Hurricane Forecasting: a novel multimodal machine learning framework. Weather and Forecasting, 37(6), 817–831. 

[31] Bryan Lim and Stefan Zohren. Time Series forecasting with deep learning: a survey. Philosophical Transactions of the Royal Society, 2021.

[32] Ashish Vaswani, Noam Shazeer, Niki Parmar, Jakob Uszkoreit, Llion Jones, Aidan N Gomez, Łukasz Kaiser, et al. Attention is all you need. In NeurIPS, 2017.

[33] George Zerveas, Srideepika Jayaraman, Dhaval Patel, Anuradha Bhamidipaty, and Carsten Eickhoff. A transformer-based framework for multivariate time series representation learning. In KDD, 2021

[34] Chao-Han Huck Yang, Yun-Yun Tsai, and PinYu Chen. Voice2series: Reprogramming acoustic models for time series classification. In ICML, 2021.

[35] Jiehui Xu, Haixu Wu, Jianmin Wang, and Mingsheng Long. Anomaly Transformer: Time series anomaly detection with association discrepancy. In ICLR, 2022.

[36] Chulhee Yun, Srinadh Bhojanapalli, Ankit Singh Rawat, Sashank J. Reddi, et al. Are transformers universal approximators of sequence-to-sequence functions? In ICLR, 2020.

[37] Fout, A., Byrd, J., Shariat, B., and Ben-Hur, A. Protein interface prediction using graph convolutional networks. In NeurIPS, pp. 6530–6539, 2017

[38] Guo, S., Lin, Y., Feng, N., Song, C., and Wan, H. Attention based spatial-temporal graph convolutional networks for traffic flow forecasting. In AAAI, 2019.

[39] Qiu, J., Tang, J., Ma, H., Dong, Y., Wang, K., and Tang, J. Deepinf: Social influence prediction with deep learning. In KDD, pp. 2110–2119. ACM, 2018

[40] Girshick, Ross. "Fast r-cnn." Proceedings of the IEEE international conference on computer vision. 2015.

[41] https://www.ready.gov/sites/default/files/2021-11/are-you-ready-guide.pdf






\newpage

\section{Supplementary Material}
Here, we show qualitatively how GraphTransformers outshine vanilla Transformers in the next location prediction task. The groundtruth latitude-longitude along with the predicted location for the both the models for randomly sampled examples in the test split are shown as follows.
\begin{figure}[!htb]
  \centering
  \includegraphics[scale=0.76]{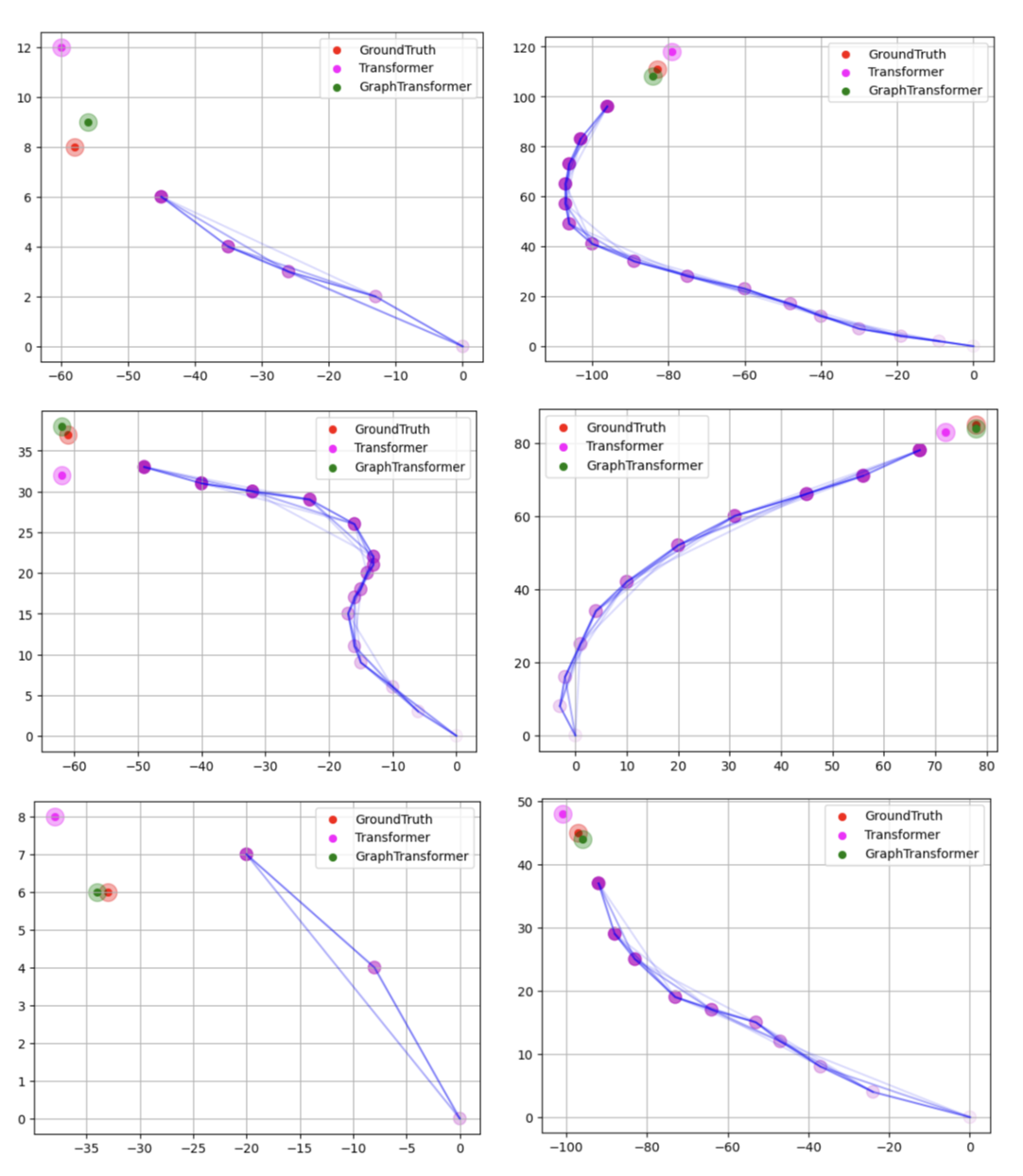}
  \caption{Qualtitative results shown on randomly sampled input trajectories from test set showcasing the superiority of the GraphTransformer network. The purple dots with blue lines connecting show the input trajectory along with their edge strength.}
\end{figure} 

\end{document}